%% file: root.tex
\documentclass[letterpaper, 10 pt, conference]{ieeeconf}  

\IEEEoverridecommandlockouts                              

\usepackage{graphicx}
\usepackage{amsmath}
\usepackage{amssymb}
\usepackage{hyperref}
\usepackage{caption}
\usepackage{comment}
\usepackage{xcolor}
\DeclareMathOperator{\Var}{Var}
\DeclareMathOperator*{\argmin}{arg\,min}
\DeclareMathOperator*{\argmax}{arg\,max}

\title{\LARGE \bf
Uncertainty-driven Planner for Exploration and Navigation
\thanks{Research was sponsored by the Army Research Office and was accomplished under Grant Number W911NF-20-1-0080. The views and conclusions contained in this document are those of the authors and should not be interpreted as representing the official policies, either expressed or implied, of the Army Research Office or the U.S. Government. The U.S. Government is authorized to reproduce and distribute reprints for Government purposes notwithstanding any copyright notation herein. Further support was provided by the following grants: NSF IIS 1703319, NSF MRI 1626008, NSF TRIPODS 1934960, NSF CPS 2038873, ARL DCIST CRA W911NF-17-2-0181, ONR N00014-17-1-2093, the DARPA-SRC C-BRIC, CAREER award ECCS-2045834, and a Google Research Scholar award.
}
}

\author{Georgios Georgakis$^{1}$, Bernadette Bucher$^{1}$, Anton Arapin$^{2}$, Karl Schmeckpeper$^{1}$, \\ Nikolai Matni$^{1}$, and Kostas Daniilidis$^{1}$
\thanks{$^{1}$GRASP Laboratory, Department of Computer and Information Science, University of Pennsylvania, Philadelphia, PA 19104. 
        {\tt\small ggeorgak@seas.upenn.edu}}%
\thanks{$^{2}$Department of Computer Science, The University of Chicago, Chicago, IL, 60637. 
        {\tt\small aarapin@uchicago.edu}}%
}

\begin{document}

\maketitle
\thispagestyle{empty}
\pagestyle{empty}

\begin{abstract}
We consider the problems of exploration and point-goal navigation in previously unseen environments, where the spatial complexity of indoor scenes and partial observability constitute these tasks challenging. We argue that learning occupancy priors over indoor maps provides significant advantages towards addressing these problems. To this end, we present a novel planning framework that first learns to generate occupancy maps beyond the field-of-view of the agent, and second leverages the model uncertainty over the generated areas to formulate path selection policies for each task of interest. For point-goal navigation the policy chooses paths with an upper confidence bound policy for efficient and traversable paths, while for exploration the policy maximizes model uncertainty over candidate paths. We perform experiments in the visually realistic environments of Matterport3D using the Habitat simulator and demonstrate: 1) Improved results on exploration and map quality metrics over competitive methods, and 2) The effectiveness of our planning module when paired with the state-of-the-art DD-PPO method for the point-goal navigation task.


\end{abstract}

\input{tex/introduction}

\input{tex/related_work}

\input{tex/approach}
\input{tex/experiments}

\input{tex/conclusion}

\bibliographystyle{./IEEEtran} 
\bibliography{./IEEEexample}

\end{document}

%% file: tex/introduction.tex
\section{Introduction}
A major prerequisite towards true autonomy is the ability to navigate and explore novel environments. This problem is usually studied in the context of specific tasks such as reaching a specified point goal~\cite{anderson2018evaluation}, finding a semantic target~\cite{batra2020objectnav}, or covering as much area as possible while building a map. 
Each of these tasks has its own idiosyncrasies, but all of them represent examples where one must often reason beyond what is currently observed and incorporate the uncertainty over the inferred information into the decision making process.
For example, in point-goal navigation it is important to predict whether a certain path can lead to a dead-end. Likewise, in exploration strong confidence over a particular region's representation may prompt the agent to visit new areas of the map.

We investigate the tasks of point-goal navigation and exploration, and propose a planning module that leverages contextual occupancy priors. These priors are learned by a map predictor module that is trained to estimate occupancy values outside the field-of-view of the agent. Using the epistemic (model) uncertainty associated with these predictions we define objectives for path selection for each task of interest.
Earlier work in this field focused mainly on learning how to actively control the agent for the purpose of reducing the uncertainty over the map~\cite{cadena2016past} (Active SLAM), without considering navigation tasks in the process, while methods that did consider navigation often operated in relatively simple environments of artificially placed cylindrical obstacles~\cite{melchior2007particle,ok2013path}.

With the recent introduction of realistic and visually complex environments serving as navigation benchmarks~\cite{savva2019habitat,xia2018gibson},
the focus shifted on learning-based end-to-end approaches~\cite{zhu2017target,chen2019learning,wijmans2019dd}. While end-to-end formulations that map pixels directly to actions are attractive in terms of their simplicity, they require very large quantities of training data. For instance, DD-PPO~\cite{wijmans2019dd} needs 2.5 billion frames of experience to reach its state-of-the-art performance on Gibson~\cite{xia2018gibson}.
On the other hand, modular approaches~\cite{gupta2017cognitive,ramakrishnan2020occupancy,chaplot2020learning} are able to encode prior information into explicit map representations and are thus much more sample efficient. 
Our method falls into the latter category, but differs from other approaches by its use of the uncertainty over predictions outside the field-of-view of the agent during the planning stage. In contrast to~\cite{chaplot2020learning,ramakrishnan2020occupancy} this allows our method more flexibility when defining goal selection objectives, and does not require re-training between different tasks.

\begin{figure*}[t]
    \vspace{2mm}
    \centering
    \includegraphics[width=0.85\linewidth]{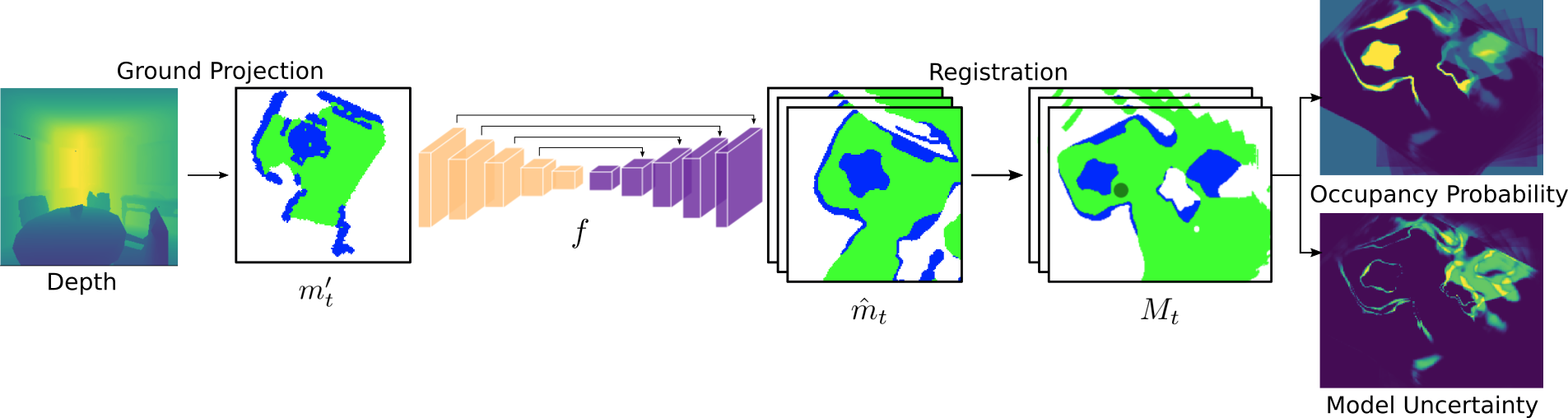}
    \caption{Occupancy map prediction (blue-occupied, green-free) and uncertainty estimation for a time-step $t$. The egocentric depth observation is first ground-projected and passed through an ensemble $f$ of encoder-decoder models that each infers information in unobserved areas ($\hat{m}_t$). Each $\hat{m}_t$ is then registered to a separate global map $M_t$. The final occupancy probabilities and model uncertainty are given by the mean and variance over the set of global maps. 
    }
    \label{fig:map_predictor}
    \vspace{-2mm}
\end{figure*}

In this paper, we introduce \textit{Uncertainty-driven Planner for Exploration and Navigation} (UPEN), in which we propose a planning algorithm that is informed by predictions over unobserved areas. Through this spatial prediction approach our model learns layout patterns that can guide a planner towards preferable paths in unknown environments. 
More specifically, we first train an ensemble of occupancy map predictor models by learning to hallucinate top-down occupancy regions from unobserved areas. 
Then, a Rapidly Exploring Random-Trees~\cite{lavalle1998rapidly} (RRT) algorithm generates a set of candidate paths. We select paths from these candidates using epistemic (model) uncertainty associated with a path traversibility estimate as measured by the disagreement of ensemble models~\cite{seung1992query, Pathak2019}, and we choose appropriate short-term goals based on the task of interest.
Our contributions are as follows:
\begin{itemize}
    \item We propose UPEN, a novel planning framework that leverages learned layout priors and formulates uncertainty-based objectives for path selection in exploration and navigation tasks.
    \item We show improved exploration results over competitive methods on the Matterport3D~\cite{chang2017matterport3d} dataset.
    \item We demonstrate the effectiveness of our planner when used to complement existing end-to-end methods on the point-goal navigation task.
\end{itemize}








%% file: tex/related_work.tex
\section{Related Work}

\paragraph{Navigation approaches} Traditional approaches to visual navigation focus on building a 3D metric map of the environment~\cite{fuentes2015visual,cadena2016past} before using that representation for any downstream navigation tasks, which does not lend itself favourably for task-driven learnable representations that can capture contextual cues. The recent introduction of large-scale indoor environments and simulators~\cite{xia2018gibson,chang2017matterport3d,savva2019habitat} has fuelled a slew of learning based methods for indoor navigation tasks~\cite{anderson2018evaluation} such as point-goal~\cite{wijmans2019dd,savva2017minos,zhao2021surprising,karkus2021differentiable,mishkin2019benchmarking}, object-goal~\cite{chaplot2020object,georgakis2021learning,georgakis2019simultaneous,mousavian2019visual,liang2020sscnav}, and image-goal~\cite{zhu2017target,chaplot2020neural,kwonvisual}. Modular approaches which incorporate explicit or learned map representations~\cite{gupta2017cognitive,chaplot2020object,georgakis2019simultaneous} have shown to outperform end-to-end methods on tasks such as object-goal, however, this is not currently the case for the point-goal~\cite{wijmans2019dd,zhao2021surprising} task. 
In our work, we demonstrate how an uncertainty-driven planning module can favourably complement DD-PPO~\cite{wijmans2019dd}, 
a competitive method on point-goal navigation,
and show increased performance in challenging episodes. 

\paragraph{Exploration methods for navigation} A considerable amount of work was also devoted to planning efficient paths during map building, generally referred to as Active SLAM~\cite{feder1999adaptive,kollar2008trajectory,carlone2014active,carrillo2012comparison,blanco2008novel,stachniss2005information}. For example, ~\cite{carlone2014active,stachniss2005information} define information gain objectives based on the estimated uncertainty over the map in order to decide future actions, while~\cite{carrillo2012comparison} investigates different uncertainty measures. Recent methods focus on learning policies for efficient exploration either through coverage~\cite{chen2019learning,chaplot2020learning,fang2019scene,zhang2017neural} or map accuracy~\cite{ramakrishnan2020occupancy} reward functions. Furthermore, several works have gone beyond traditional mapping, and sought to predict maps for unseen regions~\cite{ramakrishnan2020occupancy,narasimhan2020seeing,georgakis2021learning,liang2020sscnav,katsumata2020spcomapgan} which further increased robustness in the decision making process. Our approach leverages the uncertainty over predicted occupancy maps for unobserved areas and shows its effectiveness on exploring a novel environment.

\paragraph{Uncertainty estimation} 
To navigate in partially observed maps, uncertainty has been estimated across nodes in a path \cite{melchior2007particle, beeching2020learning}, via the marginal probability of landmarks \cite{ok2013path}, and with the variance of model predictions across predicted maps \cite{georgakis2021learning, 8793500}. 
Furthermore, uncertainty-aware mapping has been shown to be effective in unknown and highly risky environments~\cite{fan2021step,pairet2021online}.
In our work, we use uncertainty differently for exploration and point goal navigation. In exploration, we estimate uncertainty over a predicted occupancy map via the variance between models in an ensemble. This variance across the ensemble specifically estimates model (epistemic) uncertainty \cite{Gal2016Uncertainty, kendall2017uncertainties}. We select paths by maximizing epistemic uncertainty as a proxy for maximizing information gain following prior work in exploration \cite{Pathak2019, georgakis2021learning}. In point goal navigation, we compute traversability scores for candidate paths using an ensemble of map predictors and compute uncertainty with respect to these traversability scores using the variance over the scores given by each model in the ensemble. We use this uncertainty regarding path traversability to construct an upper confidence bound policy for path selection to balance exploration and exploitation in point goal navigation \cite{azar2017minimax, auer2002finite, chen2017ucb, georgakis2021learning}.




%% file: tex/approach.tex
\section{Approach}

\begin{figure*}[t]
    \vspace{2mm}
    \centering
    \includegraphics[width=0.85\linewidth]{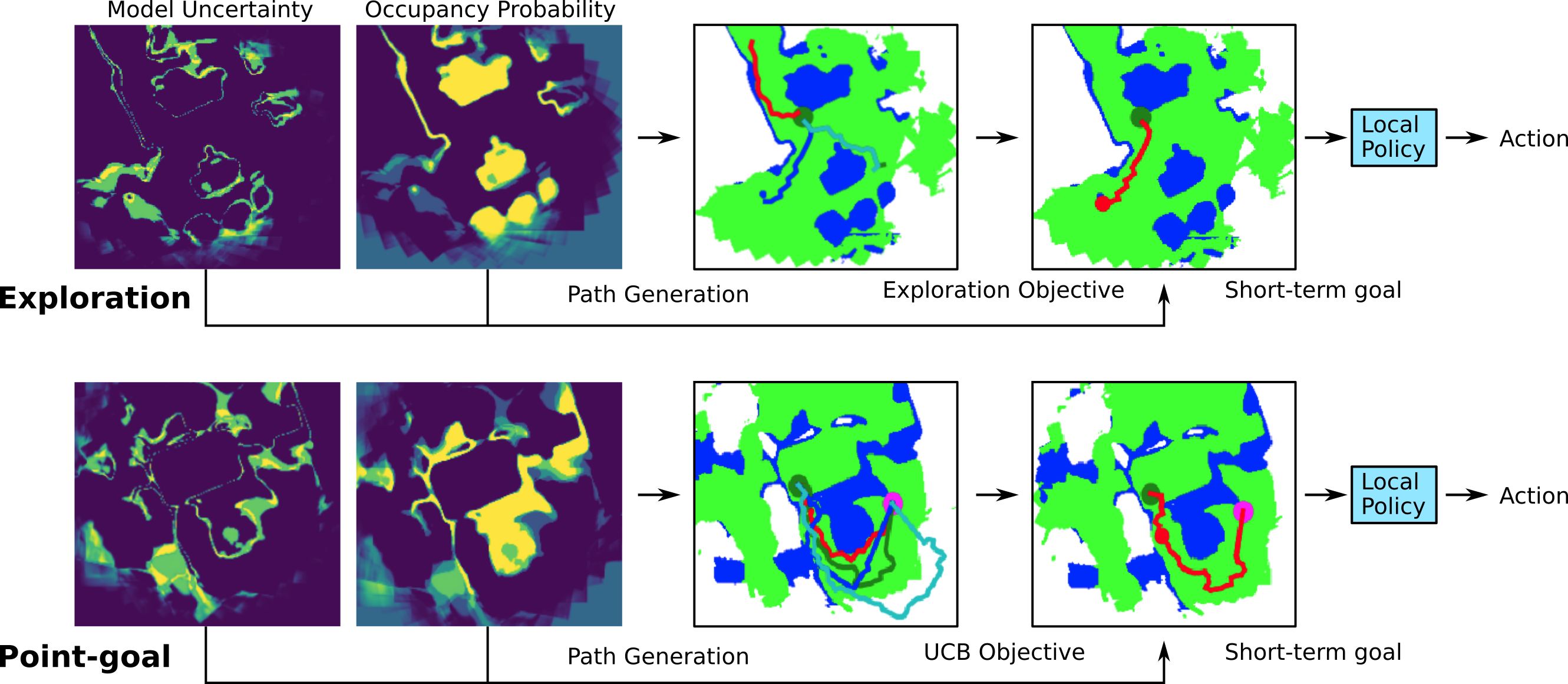}
    \caption{Examples of path selections for exploration (top row) and point-goal navigation (bottom-row) tasks. Given the model uncertainty and occupancy probabilities we first generate a set of paths which are evaluated either with an exploration objective (section~\ref{subsec:exploration_approach}) or an upper confidence bound objective (section~\ref{subsec:pointgoal_approach}). The agent position is denoted as a dark green dot, the goal is shown as magenta, and red dots signify short-term goals.}
    \label{fig:system_policy}
    \vspace{-2mm}
\end{figure*}

We present an uncertainty-driven planning module for exploration and point-goal navigation tasks, which benefits from a learned occupancy map predictor module. 
Our approach takes as input the agent's egocentric depth observation and learns to predict regions of the occupancy map that are outside of the agent's field-of-view. Then it uses the uncertainty over those predictions to decide on a set of candidate paths generated by RRT. We define a separate policy to select a short-term goal along a path for each task of interest. In exploration we maximize uncertainty over the candidate paths, while for point-goal navigation we choose paths with an upper confidence bound policy for efficient and traversable paths.
Finally, a local policy (DD-PPO~\cite{wijmans2019dd}) predicts navigation actions to reach the short-term goal. 

\subsection{Occupancy Map Prediction}
The first component in our planning module aims to capture layout priors in indoor environments. Such information can lead to a more intelligent decision making process for a downstream navigation task. Following the recent success of ~\cite{ramakrishnan2020occupancy,georgakis2021learning} we formulate the occupancy map prediction as a semantic segmentation problem. Our model takes as input a depth image $D_t$ at time-step $t$ which is ground projected to an egocentric grid $m'_t \in \mathbb{R}^{|C| \times h \times w}$, where $C$ is the set of classes containing $unknown$, $occupied$, and $free$, and $h$, $w$ are the dimensions of the local grid. The ground projection is carried out by first using the camera intrinsic parameters to unproject $D_t$ to a 3D point cloud and then map each 3D point to the $h \times w$ grid coordinates: $x' = \lfloor \frac{x}{r} \rfloor + \frac{w-1}{2}$, $z' = \lfloor \frac{z}{r} \rfloor + \frac{h-1}{2}$, where $x'$, $z'$ are the grid coordinates, $x$, $z$ are the 3D points, and $r$ is the grid cell size. Since the agent has a limited field of view, $m'_t$ represents a local incomplete top-down occupancy grid of the area surrounding the agent. Our objective is to predict the missing values and produce the complete local occupancy map $\hat{m}_t \in \mathbb{R}^{|C| \times h \times w}$. To do so, we pass $m'_t$ through an encoder-decoder UNet~\cite{ronneberger2015u} model $f$ that outputs a prediction for each grid location over the set of classes $C$.
The model $f$ is trained with a pixel-wise cross-entropy loss:
\begin{equation}
    L = - \frac{1}{K} \sum_k^K \sum_c^{C} m_{k,c} \log \hat{m}_{k,c}
\end{equation}
where $K = h \times w$ corresponds to the number of cells in the local grid and $m_{k,c}$ is the ground-truth label for pixel k. The ground-truth occupancy is generated by ground-projecting the available semantic information of the 3D scenes. To ensure diversity in the training examples, we sample training pairs across shortest paths between two randomly selected locations in a scene, where $m'_t$ can contain a variable number of ground-projected depth images. Unlike~\cite{ramakrishnan2020occupancy} we do not use the RGB images during training, as we have found that the aforementioned sampling strategy is sufficient for the model to converge. This enables us to define a smaller and less memory intensive model $f$.

During a navigation episode, we maintain a global map $M_t \in \mathbb{R}^{|C| \times H \times W}$. Since $f$ predicts a probability distribution over the classes for each grid location, we register $\hat{m}_t$ by updating $M_t$
using Bayes Theorem. The global map $M_t$ is initialized with a uniform prior probability distribution across all classes. 


\subsection{Exploration Policy}\label{subsec:exploration_approach} The main goal of exploration task is to maximize map coverage which requires navigating to new map regions around obstacles. To this end, we propose selecting paths using uncertainty of our map predictions as an objective in our planning algorithm. We are explicitly minimizing map uncertainty by collecting observations to improve the predicted global map $M_t$. Implicitly map coverage is maximized by minimizing map uncertainty because high coverage is required for predicting an accurate map with low uncertainty. 

We use the epistemic (model) uncertainty as an objective for exploration \cite{kendall2017uncertainties, Gal2016Uncertainty, Pathak2019, georgakis2021learning}. In order to estimate epistemic uncertainty, we construct $f$ as an ensemble of $N$ occupancy prediction models defined over the parameters $\lbrace \theta_1, ..., \theta_N \rbrace$. Variance between models in the ensemble comes from different random weight initializations in each network \cite{Pathak2019}.
Our model estimates the true probability distribution $P(m_t | m_t')$ by averaging over sampled model weights, $P(m_t | m_t') \approx \mathbb{E}_{\theta \sim q(\theta)}f(m_t'; \theta) \approx \frac{1}{N}\sum_{i=1}^N f(m_t'; \theta_i)$ where the parameters $\theta$ are random variables sampled from the distribution $q(\theta)$\cite{lakshminarayanan2016simple, gal2017deep}. Then, following prior work~\cite{seung1992query,Pathak2019,georgakis2021learning}, the epistemic uncertainty can be approximated from the variance between the outputs of the models in the ensemble, $\Var f(m_t'; \theta)$.

For path planning during exploration, our proposed objective can be used with any planner which generates a set $S$ of candidate paths. Each path $s \in S$ can be expressed as a subset of grid locations in our map. Each of these grid locations $k$ has an associated uncertainty estimate given by the variance between model predictions in our ensemble. We specify this uncertainty map as $u_{k} := \Var f(m_t'; \theta) \in \mathbb{R}^{1\times h \times w}$. We use this map to score each path $s$ and construct the objective 
\begin{equation}\label{eq:explore}
    \argmax_{s \in S} \frac{1}{|s|}\sum_{k \in s} u_k
\end{equation}
which selects the path with the maximum average epistemic uncertainty on the traversed grid.

In this work, we incorporate our uncertainty-based objective in RRT to plan to explore. We expand RRT for a set number of iterations, which generates candidate paths in random directions. We select between these paths using our objective from equation~\ref{eq:explore}. In practice, equation~\ref{eq:explore} is evaluated over the accumulated global map $M_t$. Figure~\ref{fig:map_predictor} shows the occupancy map prediction and the uncertainty estimation process using the ensemble $f$, while Figure~\ref{fig:system_policy} (top row) shows an example of path selection using the exploration objective.


\subsection{Point-goal Policy}\label{subsec:pointgoal_approach} In the problem of point-goal navigation, the objective is to efficiently navigate past obstacles to a given goal location from a starting position. We again use RRT as a planner which generates a set of paths $S$ between the agent's current location and the goal location. Thus, the primary objective when we select a path from these candidates to traverse is for the path to be unobstructed. Given a predicted occupancy map from model $i$ in our ensemble and a candidate path $s \in S$ generated by our planner, we evaluate whether or not the path is obstructed by taking the maximum probability of occupancy in any grid cell $k$ along each path. Specifically, 
\begin{equation}
    p_{i, s} = \max_{k \in s} \left(\hat{m}_{k, occ}^i|_{k \in s} \right)
\end{equation}
where $\hat{m}^i_{k, occ}|_{k \in s}$ is the map of occupancy probabilities defined on the subset of grid cells $k \in s$ predicted by model $i$ in the ensemble $f$.
Choosing the path $s \in S$ by minimizing $p_{i, s}$ chooses the path we think most likely to be unobstructed. We can minimize this likelihood by selecting $\argmin_{s \in S} \mu_s$ where $\mu_s := \frac{1}{N} \sum_{i=1}^N p_{i, s}$. However, we note that there may be multiple unobstructed candidate paths generated by our planner. We differentiate between these in our selection by adding a term $d_s$ to our objective to incentivize selecting shorter paths.
Furthermore, as an agent navigates to a goal, it makes map predictions using its accumulated observations along the way. Therefore, to improve navigation performance we can incorporate an exploration component in our navigation objective to incentivize choosing paths where it can gain the most information regarding efficient traversability. 



We estimate uncertainty associated with efficient traversability of a particular path $s$ for our exploration objective. Since there is zero uncertainty associated with path lengths $d_s$, we design our exploration objective to maximize information gain for path traversability. We denote $P_{s_{NT}}(m_t|m_t')$ as the probability the path $s$ is not traversable ($NT$) estimated by $\mu_s$. We recall that $\mu_s$ is computed by averaging traversability scores over an ensemble of models. We compute the variance of these scores $\Var_{i \in N} p_{i, s}$ to estimate uncertainty of our model approximating $P_{s_{NT}}(m_t|m_t')$. 


We combine exploration and exploitation in our full objective using an upper confidence bound policy \cite{auer2002finite,azar2017minimax,chen2017ucb,georgakis2021learning}. Our objective for efficient traversable paths is specified as
\begin{equation}
    \argmin_{s \in S} P_{s_{NT}}(m_t|m_t') + d_s
\end{equation}
and can be reconstructed as a maximization problem $\argmax_{s \in S} - P_{s_{NT}}(m_t|m_t') - d_s$. We denote $\sigma_s := \sqrt{\Var_{i \in N} p_{i, s}}$ and observe the upper bound
\begin{equation}
    -P_{s_{NT}}(m_t|m_t') - d_s \leq -\mu_s + \alpha_1 \sigma_s - d_s
\end{equation}
holds with some fixed but unknown probability where $\alpha_1$
is a constant hyperparameter.
Using our upper bound to estimate $-P_{s_{NT}}(m_t|D_t)$, our full objective function as a minimization problem is
\begin{equation} \label{eq:pointnav_objective}
    \argmin_s \mu_s - \alpha_1 \sigma_s + \alpha_2 d_s
\end{equation}
where $\alpha_2$ is a hyperparameter weighting the contribution of path length. Similarly to our exploration policy, in practice, equation~\ref{eq:pointnav_objective} is evaluated over the accumulated global map $M_t$. Figure~\ref{fig:system_policy} (bottom row) illustrates path selection using our objective during a point-goal episode.

%% file: tex/experiments.tex
\section{Experiments}

\begin{figure*}[t]
    \vspace{2mm}
    \centering
    \includegraphics[width=0.9\linewidth]{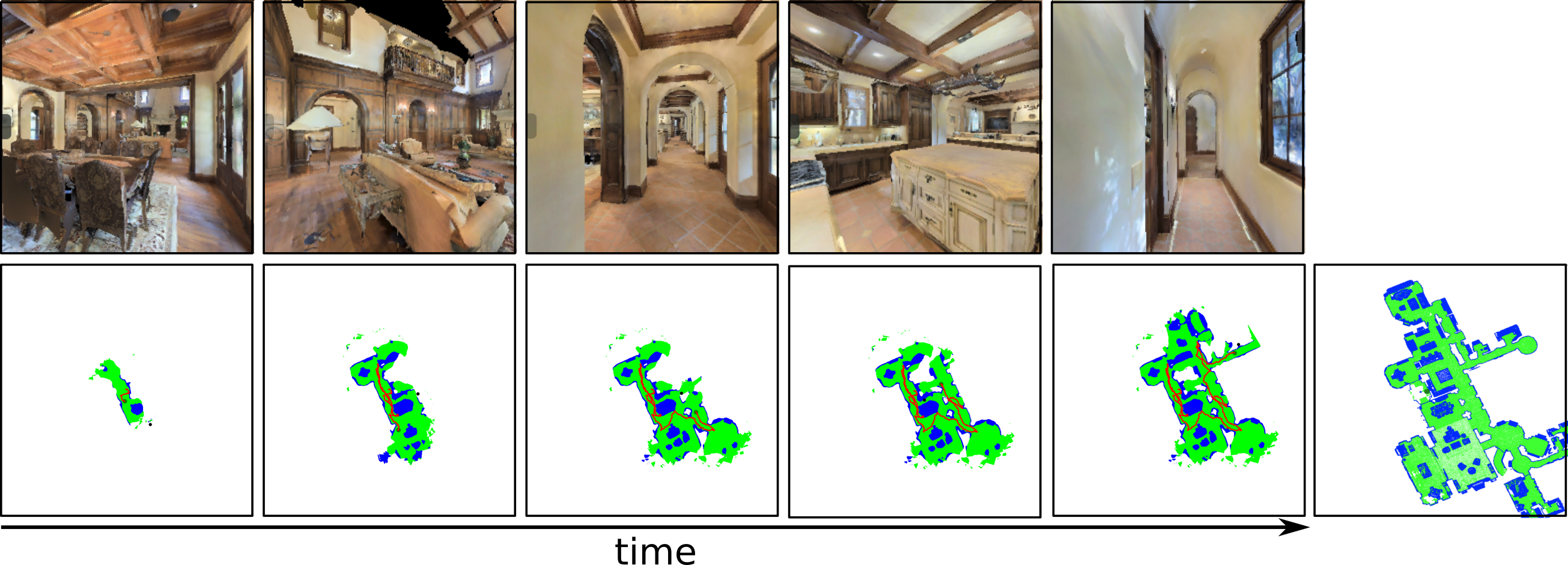}
    \caption{Exploration example with T=1000 showing the trajectory followed by our agent (red line). The top row shows RGB images observed by the agent. The ground-truth map is visualized in the bottom right corner.}
    \label{fig:exploration_examples}
    \vspace{-2mm}
\end{figure*}

\begin{table*}
\parbox{0.55\linewidth}{
    \centering
    \scalebox{0.92}{
    \begin{tabular}{|l|c|c|c|c|c|c|c|c|}
    \hline
    & \multicolumn{2}{c|}{Noisy} & \multicolumn{2}{c|}{Noise-free} \\
    \hline
    Method & Map Acc ($m^2$) & IoU (\%) & Map Acc ($m^2$) & IoU (\%) \\
    \hline
    ANS(depth)~\cite{ramakrishnan2020occupancy} & 72.5 & 26.0 & 85.9 & 34.0 \\
    OccAnt(depth) w/o AR ~\cite{ramakrishnan2020occupancy} & 92.7 & 29.0 & 104.7 & 38.0 \\
    OccAnt(depth)~\cite{ramakrishnan2020occupancy} & 94.1 & \textbf{33.0} & 96.5 & 35.0 \\
    FBE~\cite{yamauchi1997frontier} + DD-PPO~\cite{wijmans2019dd} & 100.9 & 28.7 & 120.2 & 44.7 \\
    UPEN + DD-PPO~\cite{wijmans2019dd} & \textbf{110.3} & 25.8 & \textbf{141.6} & \textbf{45.6} \\
    \hline 
    \end{tabular}
    }
    \caption{Exploration results on MP3D test scenes evaluating map quality at T=500. 
    The ``w/o AR'' refers to the baseline that is trained without the anticipation reward in~\cite{ramakrishnan2020occupancy}.}
    \label{tab:exploration_res}
}
\hfill
\parbox{.4\linewidth}{
    \centering
    \begin{tabular}{l|c|c}
     & Cov ($m^2$) & Cov (\%) \\
    \hline
    ANS(rgb)~\cite{chaplot2020learning} & 73.28 & 52.1 \\
    FBE~\cite{yamauchi1997frontier} + DD-PPO~\cite{wijmans2019dd} & 85.3 & 53.0  \\
    UPEN + DD-PPO~\cite{wijmans2019dd} & \textbf{113.0} & \textbf{67.9} \\
    \end{tabular}
    \caption{Exploration results on MP3D test scenes evaluating area coverage at T=1000.}
    \label{tab:exploration_res2}
}
\end{table*}

Our experiments are conducted on the Matterport3D (MP3D)~\cite{chang2017matterport3d} dataset using the Habitat~\cite{savva2019habitat} simulator. We follow the standard train/val/test environments split of MP3D which contains overall 90 reconstructions of realistic indoor scenes. The splits are disjoint, therefore all evaluations are conducted in novel scenes where the occupancy map predictor model has not seen during training. Our observation space consists of $256 \times 256$ depth images, while the action space contains four actions:  \texttt{MOVE\_FORWARD} by $25cm$, \texttt{TURN\_LEFT} and \texttt{TURN\_RIGHT} by $10^\circ$ and \texttt{STOP}.

We perform two key experiments. First, we compare to other state-of-the-art methods on the task of exploration using both coverage and map accuracy metrics (sec.~\ref{subsec:exploration_experiments}). Second we evaluate on the point-goal navigation task and demonstrate increased performance when DD-PPO~\cite{wijmans2019dd} is complemented with our planning strategy (sec.~\ref{subsec:pointgoal_experiments}).

\subsection{Implementation Details}
The Unet~\cite{ronneberger2015u} model used for the occupancy map prediction has four encoder and four decoder convolutional blocks with skip connections and it is combined with a ResNet18~\cite{he2016deep} for feature extraction. We use Pytorch~\cite{paszke2017automatic} and train using the Adam optimizer with a learning rate of 0.0002. The grid dimensions are $h=w=160$ for local, and $H=W=768$ for global, while each cell in the grid is $5cm \times 5cm$. For the path generation process, we run the RRT every 30 navigation steps for exploration and 20 for point-goal. The RRT is set to generate a maximum of 10 paths every run, with a goal sampling rate of 20\%. Finally, the RRT expands new nodes with a distance of 5 pixels at a time. A single step in a navigation episode requires 0.37s on average that includes map prediction and registration, planning using RRT, and DD-PPO. The timing was performed on a laptop using i7 CPU @ 2.20GHz and a GTX1060 GPU.
All experiments are with ensemble size of 4. We provide code and trained models: \url{https://github.com/ggeorgak11/UPEN}.

\subsection{Exploration} \label{subsec:exploration_experiments}
The setup from~\cite{ramakrishnan2020occupancy} is followed for this experiment, where the objective is to cover as much area as possible given a limited time budget $T=1000$. Unless stated otherwise, the evaluation is conducted with simulated noise following the noise models from~\cite{chaplot2020learning,ramakrishnan2020occupancy}.
We use the following metrics: 1) \textit{Map Accuracy ($m^2$)}: as defined in~\cite{ramakrishnan2020occupancy} the area in the predicted occupancy map that matches the ground-truth map. 2) \textit{IoU (\%)}: the intersection over union of the predicted map and the ground-truth. 3) \textit{Cov ($m^2$)}: the actual area covered by the agent. 4) \textit{Cov (\%)}: ratio of covered area to max scene coverage. We note that the two coverage metrics are computed on a map containing only ground-projections of depth observations. 
Our method is validated against the competitive approaches of Occupancy Anticipation~\cite{ramakrishnan2020occupancy} (\textit{OccAnt}) and Active Neural SLAM~\cite{chaplot2020learning} (\textit{ANS}), which are modular approaches with mapper components. Both use reinforcement learning to train goal selection policies optimized over map accuracy and coverage respectively. Furthermore, we compare against the classical method of Frontier-based Exploration~\cite{yamauchi1997frontier} (\textit{FBE}). Since both UPEN and FBE are combined with DD-PPO and use the same predicted maps, this comparison directly validates our exploration objective.

We report two key results. First,
in Table~\ref{tab:exploration_res} our method outperforms all baselines in the noise-free case in both \textit{Map Accuracy} and \textit{IoU}. In fact, we show $21.4m^2$ and $36.9m^2$ improvement over \textit{FBE} and \textit{OccAnt} respectively on the \textit{Map Accuracy} metric. In the noisy case even though we still surpass all baselines on \textit{Map Accuracy}, our performance drops significantly in both metrics. In addition, the \textit{Map Accuracy} increasing while \textit{IoU} drops is attributed to increased map coverage with reduced accuracy.
This is not surprising since unlike \textit{OccAnt} and \textit{Neural SLAM} we 
are not using a pose estimator.
Second, in Table~\ref{tab:exploration_res2} we demonstrate superior performance on coverage metrics with a margin of $27.7m^2$ from \textit{FBE} and $39.7m^2$ from \textit{ANS}. This suggests that our method is more efficient when exploring a novel scene, thus validating our uncertainty-based exploration policy. Figure~\ref{fig:exploration_examples} shows an exploration episode. 

\begin{table*}
\vspace{2mm}
\begin{center}
\begin{tabular}{|l|c|c|c|c|c|c|}
\hline
Dataset & \multicolumn{2}{c|}{MP3D Val} & \multicolumn{2}{c|}{MP3D Test} & \multicolumn{2}{c|}{MP3D Val-Hard} \\
\hline
Method & Success (\%) & SPL (\%) & Success (\%) & SPL (\%) & Success (\%) & SPL (\%)  \\
\hline
DD-PPO~\cite{wijmans2019dd} & 47.8 & \textbf{38.7}  & 37.3 & 30.2 & 38.0 & 28.1 \\
UPEN-Occ + DD-PPO~\cite{wijmans2019dd} & 43.8 & 30.2 & 36.3 & 25.3 & 42.3 & 26.9 \\
UPEN-Greedy + DD-PPO~\cite{wijmans2019dd} & 48.9 & 36.0 & 37.5 & 28.1 & 43.0 & 28.8 \\
UPEN + DD-PPO~\cite{wijmans2019dd} & \textbf{49.8} & 36.9 & \textbf{40.8} & \textbf{30.7} & \textbf{45.7} & \textbf{31.6} \\ 
\hline
\end{tabular}
\end{center}
\caption{Point-goal navigation results of our method against the vanilla DD-PPO\cite{wijmans2019dd}. ``Occ'' signifies a policy that uses only occupancy predictions, while ``Greedy'' refers to a policy taking into consideration path length without uncertainty.}  
\label{tab:navigation_res}
\vspace{-2mm}
\end{table*}

\begin{table}[]
    \centering
    \begin{tabular}{l|c|c|c}
         & Avg GD (m) & Avg GEDR & Min GEDR\\
         \hline
       Gibson Val  & 5.88 & 1.37 & 1.00 \\
       MP3D Val & 11.14 & 1.40 & 1.00 \\ 
       MP3D Test & 13.23 & 1.42 & 1.00 \\
       MP3D Val-Hard & 8.28 & 3.19 & 2.50 \\
    \end{tabular}
    \caption{Geodesic distance (GD) and geodesic to Euclidean distance ratio (GEDR) between different evaluation sets for point-goal navigation.}
    \label{tab:episode_stats}
\end{table}

\begin{figure}[t]
    \centering
    \includegraphics[width=0.55\linewidth]{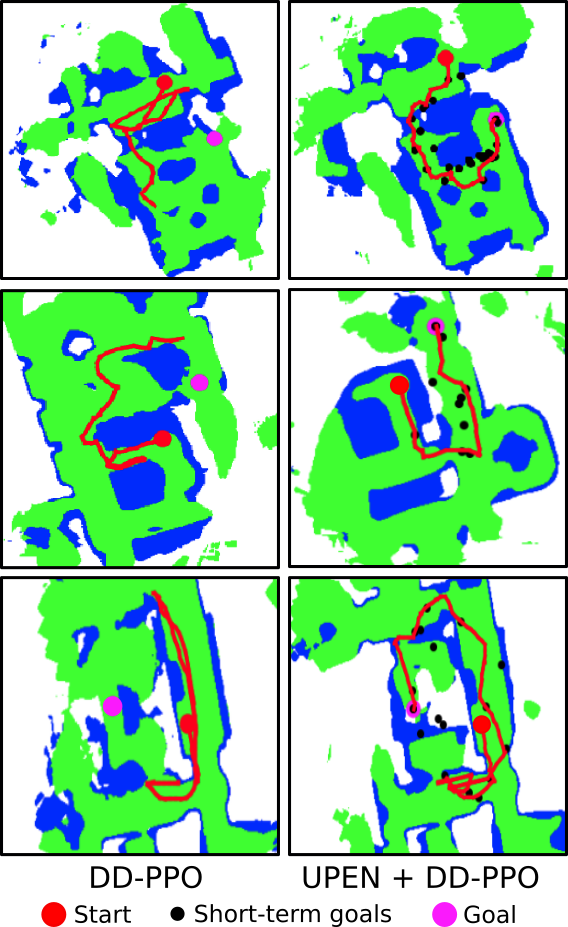}
    \caption{Point-goal navigation examples from the MP3D Val-Hard set where the vanilla DD-PPO~\cite{wijmans2019dd} fails to reach the target while our method is successful. }
    \label{fig:pointnav_examples}
\end{figure}

\subsection{Point-goal Navigation} \label{subsec:pointgoal_experiments}
We evaluate the performance of our uncertainty-driven planner when used to augment DD-PPO~\cite{wijmans2019dd} against its vanilla version. DD-PPO is currently one of the best performing methods on point-goal navigation, achieving $97\%$ SPL on the Gibson~\cite{xia2018gibson} validation set as shown in~\cite{wijmans2019dd}. We follow the point-goal task setup from~\cite{anderson2018evaluation} where given a target coordinate the agent needs to navigate to that target and stop within a $0.2m$ radius. The agent is given a time-budget of $T=500$ steps to reach the target. For evaluation we use the standard metrics~\cite{anderson2018evaluation}: \textit{Success:} percentage of successful episodes, and \textit{SPL:} success rate normalized by path length.
For this experiment we assume noise-free poses are provided by the simulator. To combine DD-PPO with our planner, we set the current short-term goal estimated by our approach as the target that DD-PPO needs to reach. For the vanilla DD-PPO we use the final target location in each test episode.

DD-PPO essentially solves Gibson point-goal navigation task so we turn our attention to MP3D where DD-PPO has lower performance due to the episodes having larger average geodesic distance (GD) to goal.
However, we noticed that the average geodesic to euclidean distance ratio (GEDR) in MP3D is still low (a GEDR of 1 means there is a straight line path between the starting position and the goal). In order to demonstrate the effectiveness of our proposed method, we generated a new evaluation set (\textit{MP3D Val-Hard}) with minimum GEDR=2.5. This created episodes which frequently involve sharp u-turns and multiple obstacles along the shortest path.
Table~\ref{tab:episode_stats} illustrates episode statistics between different evaluation sets\footnote{The Gibson val, MP3D val, and MP3D test sets were downloaded from https://github.com/facebookresearch/habitat-lab before 09/09/2021.}. In addition to \textit{MP3D Val-Hard}, we also test our method on the publicly available sets of \textit{MP3D Val} and \textit{MP3D Test}. We note that \textit{MP3D Val-Hard} was generated using the same random procedure as its publicly available counterparts.


We define two variations of our method in order to demonstrate the usefulness of our uncertainty estimation by choosing different values for the $\alpha_1$ and $\alpha_2$ parameters of Eq.~\ref{eq:pointnav_objective} from section~\ref{subsec:pointgoal_approach}. First, \textit{UPEN-Occ + DD-PPO} ($\alpha_1=0$, $\alpha_2=0$) considers only the occupancy probabilities when estimating the traversability difficulty of a path, while \textit{UPEN-Greedy + DD-PPO} ($\alpha_1=0$, $\alpha=0.5$) considers the path length and not the uncertainty.
Our default method \textit{UPEN + DD-PPO} uses $\alpha_1=0.1$ and $\alpha_2=0.5$.

The results are illustrated in Table~\ref{tab:navigation_res}. We outperform all baselines in all evaluation sets with regards to \textit{Success}. 
The largest gap in performance is observed in the \textit{MP3D Val-Hard} set which contains episodes with much higher average GEDR that the other sets.
This suggests that our method is able to follow more complicated paths by choosing short-term goals, in contrast to the vanilla DD-PPO which has to negotiate narrow passages and sharp turns only from egocentric observations. Regarding \textit{SPL}, our performance gains are not as pronounced as in \textit{Success}, 
since our policy frequently prefers paths with lower traversability difficulty in favor of shortest paths, to ensure higher success probability.

%% file: tex/conclusion.tex
\section{Conclusion}
We introduced a novel uncertainty-driven planner for exploration and navigation tasks in previously unseen environments. The planner leverages an occupancy map predictor that hallucinates map regions outside the field of view of the agent and uses its predictions to formulate uncertainty based objectives. 
Our experiments on exploration suggests that our method is more efficient in covering unknown areas.
In terms of point-goal navigation, we showed how DD-PPO~\cite{wijmans2019dd} augmented with our method outperforms its vanilla version.
This suggests that end-to-end navigation methods can benefit from employing an uncertainty-driven planner, especially in difficult episodes.